\documentclass[letterpaper]{article} 
\usepackage[preprint]{aaai2027}
\usepackage[hyphens]{url}  
\usepackage{graphicx} 
\usepackage{natbib}  
\usepackage{caption} 
\usepackage{algorithm}
\usepackage{algorithmic}

\usepackage{newfloat}
\usepackage{listings}
\DeclareCaptionStyle{ruled}{labelfont=normalfont,labelsep=colon,strut=off} 
\floatstyle{ruled}
\newfloat{listing}{tb}{lst}{}
\floatname{listing}{Listing}

\usepackage{booktabs}

\usepackage{amsmath}
\usepackage{amssymb}
\usepackage{multirow}
\usepackage{pifont}
\usepackage[table]{xcolor}

\definecolor{myred}{RGB}{180, 0, 0}
\definecolor{mypurple}{RGB}{230, 210, 250}
\definecolor{mygreen}{RGB}{0, 150, 0}
\definecolor{mypurple}{HTML}{A99BB5}

\usepackage[capitalize]{cleveref}
\crefname{section}{Sec.}{Secs.}
\Crefname{section}{Section}{Sections}
\crefname{table}{Tab.}{Tabs.}
\Crefname{table}{Table}{Tables}

\crefname{theorem}{Thm.}{Thms.}
\Crefname{theorem}{Theorem}{Theorems}
\crefname{proposition}{Prop.}{Props.}
\Crefname{proposition}{Proposition}{Propositions}
\crefname{lemma}{Lemma}{Lemmas}
\Crefname{lemma}{Lemma}{Lemmas}

\usepackage{xspace}
\makeatletter
\DeclareRobustCommand\onedot{\futurelet\@let@token\@onedot}
\def\@onedot{\ifx\@let@token.\else.\null\fi\xspace}

\def\ie{\emph{i.e}\onedot}

\makeatother

\title{Certainty Is Redundant: Token Sparsification for Efficient Camouflaged Object Detection with Vision Foundation Models}
\author{
    Yuhan Gao\textsuperscript{\rm 1,2},
    Shuhao Kang\textsuperscript{\rm 1},
    Xin He\textsuperscript{\rm 3},
    Bing Li\textsuperscript{\rm 4},
    Ming-Ming Cheng\textsuperscript{\rm 1,2,5},
    Yun Liu\textsuperscript{\rm 1,2,5}\corresponding
}
\affiliations{
    \textsuperscript{\rm 1}VCIP, College of Computer Science, Nankai University\\
    \textsuperscript{\rm 2}Academy for Advanced Interdisciplinary Studies, Nankai University\\
    \textsuperscript{\rm 3}School of Computer Science and Engineering, Tianjin University of Technology\\
    \textsuperscript{\rm 4}School of Information and Communication Engineering, UESTC\\
    \textsuperscript{\rm 5}Nankai International Advanced Research Institute, Shenzhen Futian
}

\begin{document}

\maketitle

\begin{abstract}
Camouflaged object detection (COD) aims to segment objects that closely resemble their surrounding environments. Vision foundation models (VFMs) provide strong transferable representations for COD, but their large-scale architectures and full-token processing incur substantial computational overhead. To address this issue, we propose \textbf{Certainty-Aware Token Sparsification (CATS)} for efficient VFM-based COD. Rather than estimating general token importance, CATS determines whether each token still requires deeper refinement according to its foreground--background certainty. It progressively terminates the independent updates of high-certainty tokens while retaining ambiguous tokens for further reasoning, thereby shortening the active token sequence across encoder stages. Since computational redundancy does not imply informational irrelevance, we further introduce \textbf{Dual-Path Feature Compensation (DPFC)}, which separately compresses removed foreground and background tokens into compact certainty-weighted prototypes. Extensive experiments across multiple VFMs, backbone scales, COD architectures, and benchmark datasets show that our method significantly reduces computational cost with only marginal accuracy degradation, suggesting a favorable accuracy--efficiency trade-off for VFM-based COD. The code will be released.
\end{abstract}


\section{Introduction}
Camouflaged object detection (COD) aims to identify and segment targets that blend seamlessly into their surrounding environments~\cite{fan2020camouflaged}. 
This task has significant applications in wildlife conservation \cite{fan2020camouflaged}, medical diagnosis \cite{fan2020pranet}, and precision agriculture \cite{xiao2024survey,wang2024depth,rustia2020application}. 
Unlike generic object detection, COD requires models to capture subtle discrepancies in texture, color, and structural patterns to recover complete objects with indistinct foreground–background boundaries. Although existing COD methods have gradually evolved from convolutional neural networks (CNNs) to Transformer-based architectures, reliably distinguishing camouflaged structures from visually similar backgrounds remains challenging.

\begin{figure}[!t]
    \centering
    \includegraphics[width=1.0\linewidth]{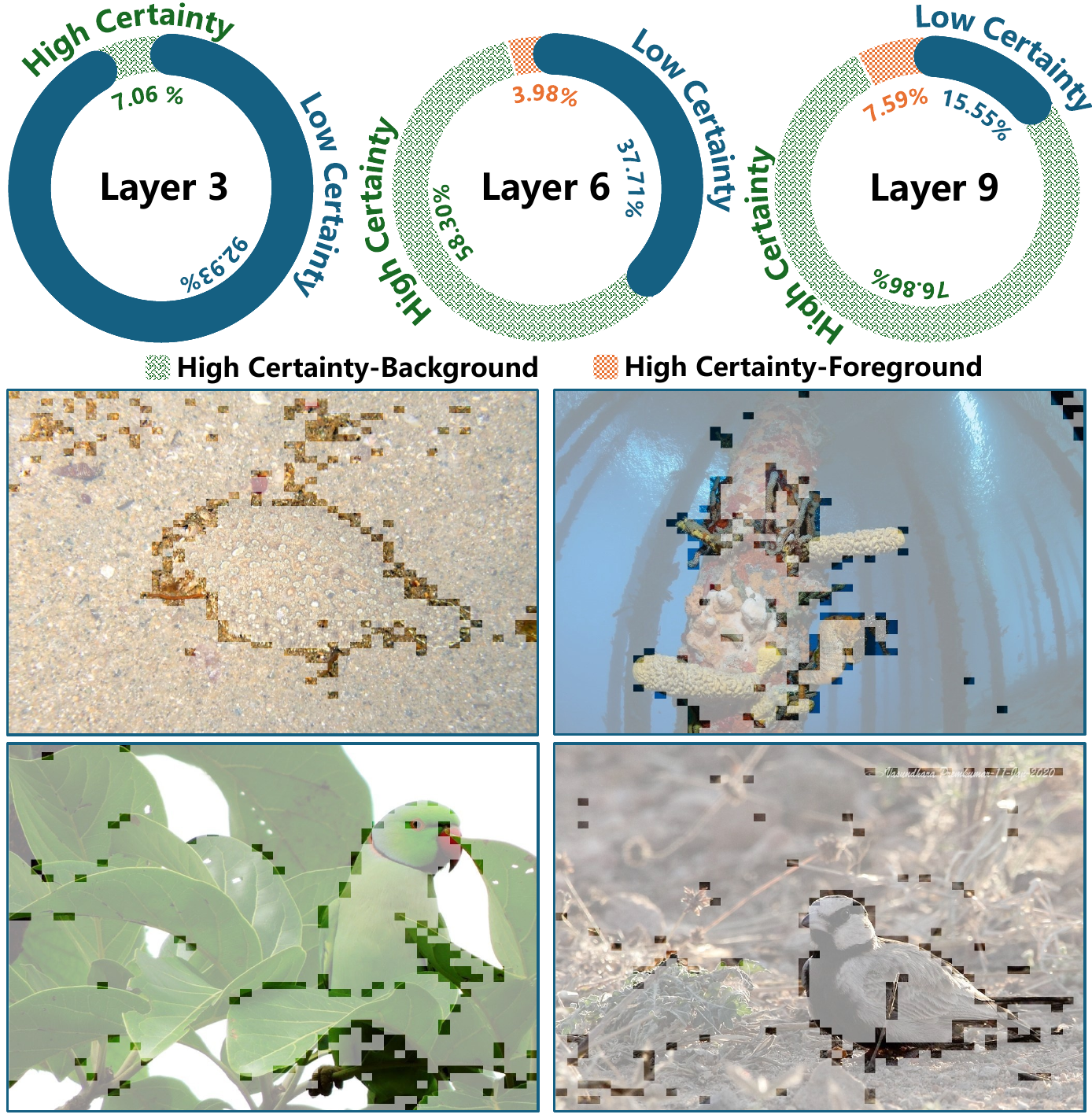}
    \caption{Layer-wise distributions of token certainty (defined in \cref{eq:certainty}), together with decision-mask visualizations. As the network deepens, more tokens exhibit high foreground or background certainty and can thus cease independent refinement. Gray regions in the bottom denote high-certainty tokens removed from further token-wise computation.}
    \label{fig:certainty}
    \vspace{-7mm}
\end{figure}

Vision foundation models (VFMs) such as MAE~\cite{he2022masked} and DINO~\cite{caron2021emerging,oquab2024dinov2,simeoni2025dinov3}, pretrained on large-scale and diverse data, provide transferable representations that encode rich semantic concepts, structural relationships, and spatial patterns. These capabilities are highly beneficial to COD, as VFMs can associate subtle object cues through global structural modeling while preserving fine-grained local representations for foreground--background discrimination. Consequently, recent studies have increasingly adopted VFMs as backbones for COD to improve feature representation and detection performance~\cite{hao2025simple,lei2025rethinking}.

Despite their strong representational capabilities, VFMs introduce substantial computational overhead when adapted to COD. All spatial tokens are repeatedly updated across multiple encoding layers, while the high-resolution inputs required to preserve subtle camouflage cues further enlarge the token sequence and exacerbate the computational burden. Since the costs of self-attention and feed-forward networks grow with the number of active tokens, reducing the token sequence provides a direct way to improve inference efficiency. Token sparsification is therefore a promising solution for the practical deployment of VFM-based COD.

Existing token sparsification methods commonly identify redundant tokens based on attention responses~\cite{liang2022not}, feature similarity~\cite{bolya2023token}, or learned importance scores~\cite{rao2021dynamicvit}, and are primarily developed for image-level recognition tasks. These criteria mainly estimate the informational importance of a token, but do not directly reflect whether a spatial token still requires further refinement. This distinction is particularly important for COD, which requires the preservation of complete object structures and fine-grained foreground–background boundaries. As illustrated in \cref{fig:certainty}, the proportion of tokens with high certainty increases as the network deepens. Once a token reaches high foreground or background certainty, repeatedly updating it across deeper layers may provide limited benefits. In contrast, uncertain tokens still require further reasoning to distinguish camouflaged objects from their visually similar surroundings. This motivates us to characterize computational redundancy from the perspective of prediction certainty, using token certainty as a task-specific indicator of the necessity for continued token-wise refinement.

In this work, we propose an adaptive \textbf{Certainty-Aware Token Sparsification (CATS)} method for efficient VFM-based COD. Rather than estimating the general informational importance of a token, CATS determines whether its foreground--background state has been sufficiently resolved and still requires further refinement. At each pruning stage, a lightweight prediction head estimates the foreground probability of each active token, from which its certainty is derived. High-certainty tokens cease independent token-wise updates, whereas ambiguous tokens remain active for deeper reasoning. Thus, CATS progressively shortens the active token sequence across encoder stages and accelerates inference.

Computational redundancy, however, does not imply informational irrelevance. Although high-certainty tokens no longer require independent refinement, they may still provide useful contextual cues for distinguishing camouflaged objects from their surroundings. We therefore introduce a \textbf{Dual-Path Feature Compensation (DPFC)} mechanism, which separately aggregates the removed background and foreground tokens into two compact prototypes. Within each group, tokens with relatively lower certainty receive larger aggregation weights, allowing the prototypes to emphasize more informative cues from the removed regions. DPFC thus preserves structured foreground and background context for subsequent computation with limited additional cost.

We evaluate the proposed method across multiple VFMs, backbone scales, and state-of-the-art (SoTA) COD baselines. Experimental results show that our method consistently reduces computational cost and improves inference speed with only marginal accuracy degradation, demonstrating a favorable accuracy--efficiency trade-off for VFM-based COD.

Our contributions are summarized as follows:
\begin{itemize}
    \item We formulate token redundancy in VFM-based COD from the perspective of continued refinement necessity and propose \textbf{CATS}, which progressively terminates the independent updates of certain tokens while retaining ambiguous tokens for deeper reasoning.
    \item We introduce \textbf{DPFC} to compress removed foreground and background tokens into certainty-weighted prototypes, preserving their contextual information for subsequent computation with limited additional cost.
    \item Extensive experiments across multiple VFMs, backbone scales, COD architectures, and benchmark datasets demonstrate that our method consistently reduces computational cost with only marginal accuracy degradation.
\end{itemize}

\section{Related Work}
\subsection{VFM-based COD}
COD has evolved from CNNs to Transformer-based architectures. SINet~\cite{fan2020camouflaged} introduced a search and identification paradigm, while FSPNet~\cite{huang2023feature} employed a feature shrinkage pyramid to progressively recover camouflaged structures. More recently, VFMs have provided transferable semantic and spatial representations learned beyond COD-specific training.

VFM-based COD studies have explored several complementary directions. SENet~\cite{hao2025simple} adapts MAE~\cite{he2022masked}, a self-supervised model pretrained through masked image modeling, to COD. DINO-based methods, including FOCUS~\cite{you2025focus}, EASE~\cite{du2025shift}, and RISE~\cite{du2025beyond}, exploit self-supervised representations for unified foreground segmentation and annotation-free camouflage discovery. SAM-based approaches adapt
promptable segmentation through dual-stream adapters \cite{liu2025improving}, multimodal fusion \cite{ren2025multimodal}, or automatic prompt refinement \cite{chen2025enhancing}, while Controllable-LPMoE \cite{sun2025controllable} injects lightweight local priors into frozen VFMs. Together, these methods demonstrate the applicability of VFMs to supervised and unsupervised COD, as well as multimodal settings.

Recent work further extends this trend. DepthSAM \cite{han2026beyond} adapts a monocular depth foundation model to camouflage-specific geometric reasoning, while UCOD-MKD \cite{chen2026beyond} and DSS~\cite{yang2026discover} explore unsupervised and zero-shot COD. DINOv3 \cite{simeoni2025dinov3} has also begun to appear in recent COD-related systems, such as GFR-SAM \cite{yang2026gfrsam} for training-free referring COD.
Nevertheless, existing studies mainly focus on representation transfer, task adaptation, or prompting, whereas inference-time token redundancy within VFM encoders remains underexplored. Our work addresses this gap through COD-specific token sparsification.

\begin{figure*}[!t]
    \centering
    \includegraphics[width=.9\textwidth]{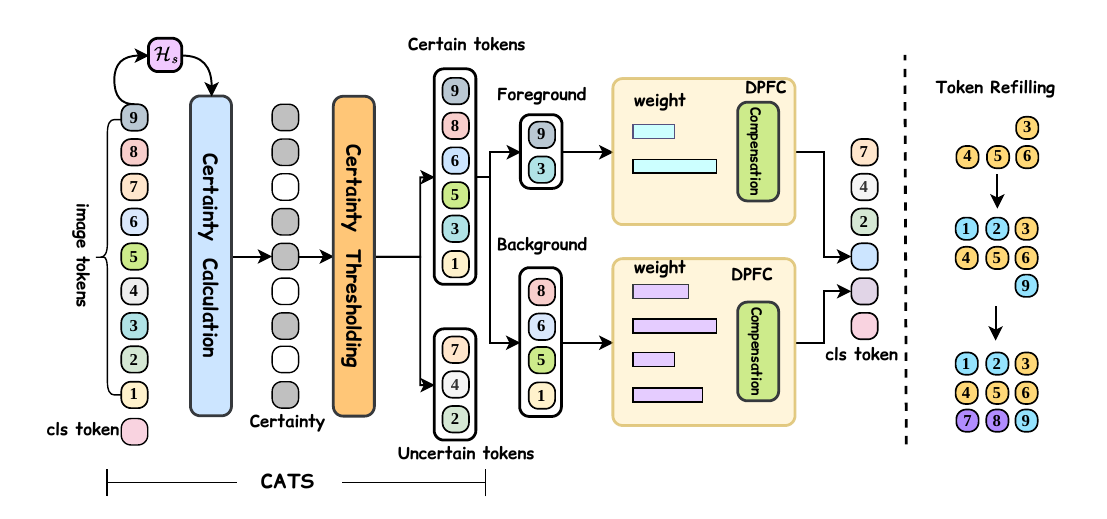}
    \caption{Overview of the proposed framework. CATS identifies high-certainty tokens as computationally redundant and removes them from subsequent token-wise computation, while retaining uncertain tokens for deeper refinement. DPFC separately aggregates the removed foreground and background tokens into two certainty-weighted prototypes to preserve useful contextual information. Finally, token refilling restores the complete spatial token layout for COD prediction.}
    \label{fig:overview}
    \vspace{-5mm}
\end{figure*}

\subsection{Token Sparsification}
Token sparsification accelerates ViT \cite{dosovitskiy2021image} by shortening the token sequence. DynamicViT~\cite{rao2021dynamicvit} progressively prunes tokens
using learned importance predictors, EViT~\cite{liang2022not} reorganizes inattentive tokens according to class-token attention, and ToMe~\cite{bolya2023token} merges similar tokens through lightweight matching. These methods are primarily developed for image recognition, whose global prediction objective does not explicitly require preserving fine-grained spatial structures. Consequently, their generic importance or similarity criteria may not transfer directly to COD.

Some studies have extended token sparsification to pixel-wise prediction.
DToP~\cite{tang2023dynamic} early-exits easy tokens in semantic segmentation and treats their intermediate predictions as final, which may be unsuitable for COD because these tokens can still provide useful context.
SViT~\cite{liu2024revisiting} preserves pruned tokens and adopts image-adaptive pruning for detection and instance segmentation. 
ALGM~\cite{norouzi2024algm} performs local-to-global token merging. 
Token Cropr~\cite{bergner2025token} learns task-relevant token selection with auxiliary heads.
DAR-TR-PEFT~\cite{lei2025rethinking} reduces token computation while preserving complete token sequences. 
However, these methods do not explicitly model the COD-specific relationship between foreground--background certainty and continued refinement. CATS instead stops independent updates for certain tokens while retaining uncertain ones for deeper reasoning. Unlike generic token reorganization or merging methods~\cite{liang2022not,bolya2023token}, DPFC further separates removed foreground and background tokens and constructs two certainty-weighted context prototypes.

\section{Methodology}
\label{sec:method}
Our framework for efficient VFM-based COD is shown in \cref{fig:overview}. First, we describe the CATS strategy to efficiently remove certain tokens while preserving critical low-certainty ones. Then, we introduce our DPFC mechanism for mitigating information loss and leveraging useful information from the dropped region. Specific details are provided below.


\subsection{Certainty-Aware Token Sparsification}
\label{sec:prune}
COD requires meticulous discrimination between foreground regions and visually similar backgrounds. Therefore, the calculation should focus more on ambiguous regions where still remain uncertain of its prediction state. As for the confidently determined tokens with high certainty, removing them from subsequent layers' computing is efficiency friendly and safe for accuracy. To reduce the active token sequence length of VFMs encoders for COD, we introduce an adaptive CATS strategy.

\begin{figure*}[!t]
    \centering 
    \includegraphics[width=0.9\textwidth]{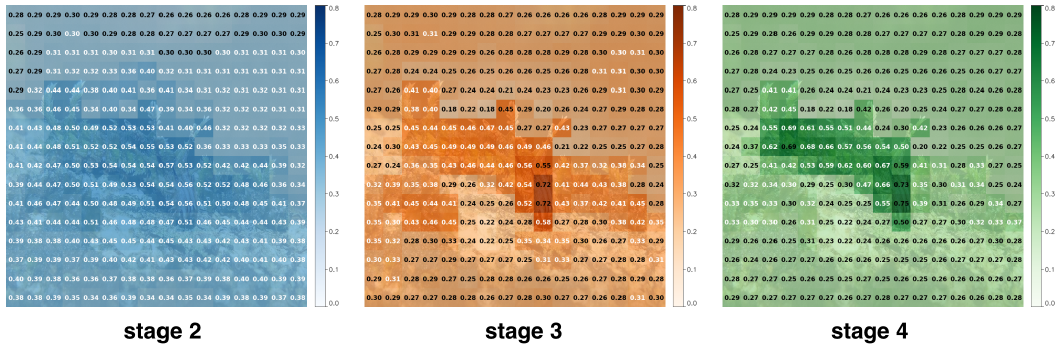} 
    \caption{Certainty heatmaps for three pruning stages. Darker values denote pruned tokens and brighter values indicate retained tokens. With progressive pruning, retained tokens gradually concentrate around object boundaries.}
    \label{fig:heat map} 
\end{figure*}

Following ViT~\cite{dosovitskiy2021image}, the input image $\mathcal{I} \in \mathbb{R}^{H \times W}$, where $H$ and $W$ represents the height and width of $\mathcal{I}$, is first partitioned into $T = HW/P^2$ non-overlapping patches of size $P \times P$. These patches are projected into a $C$-dimensional latent space via a convolutional patch embedding layer. By prepending a learnable class token $\mathbf{x}_{cls} \in \mathbb{R}^{1 \times C }$ and adding positional embeddings $\mathbf{E}_{pos}\in \mathbb{R}^{ (T+1) \times C}$, the initial sequence is formed as
\begin{equation}
\mathbf{X}^{0} = [\mathbf{x}_{cls};\ \mathrm{PatchEmbed}(\mathcal{I})] + \mathbf{E}_{pos}, \quad \mathbf{X}^{0} \in \mathbb{R}^{ (T+1) \times C}.
\end{equation}

Then, $\mathbf{X}^{0}$ is processed through $K$ Transformer layers in Stage 1. At the beginning of each subsequent pruning stage $s\in\{2,\dots,S\}$, we introduce a lightweight prediction head $\mathcal{H}_s(\cdot)$ to evaluate the $N_{s-1}$ active patch tokens inherited from stage $s-1$. Specifically, $\mathcal{H}_s(\cdot)$ consists of a lightweight scoring head $f^s_{mask}$ followed by a sigmoid activation $\sigma$, producing the prediction probabilities $\mathbf{p}^s$ of the active tokens:
\begin{equation}
\mathbf{p}^s
=
\mathcal{H}_s\left(\mathbf{X}^{(s-1)}_{\mathrm{patch}}\right)
=
\sigma\left(
\frac{
f^s_{\mathrm{mask}}\left(\mathbf{X}^{(s-1)}_{\mathrm{patch}}\right)
}{
\tau
}
\right)
,
\end{equation}
where $\mathbf{X}^{(s-1)}_{patch}\in\mathbb{R}^{N_{s-1}\times C}$ denotes the active patch tokens inherited from stage $s-1$, and $N_{s-1}$ is the number of active patch tokens, with $N_1=T$. 
The temperature factor $\tau$ is introduced to prevent sigmoid saturation before certainty estimation. When $\tau=1$, foreground and background probabilities tend to concentrate near 1 and 0 during training, respectively, causing their derived certainty scores to collapse toward similar high values and making threshold-based token selection ineffective. We therefore set $\tau=10$ to soften the probability distribution and preserve relative certainty differences, allowing CATS to better distinguish tokens that require continued refinement from those that can cease independent updates.

At each pruning stage $s\in\{2,\dots,S\}$, we use a lower threshold $\theta_d^s$ and an upper threshold $\theta_u^s$, symmetrically placed
around $0.5$, to derive the token certainty $c_t^s$ as:
\begin{equation} \label{eq:certainty}
c_t^s=2\left|p_t^s-\frac{1}{2} \right|,\qquad
a_t^s=1-c_t^s,
\end{equation}
where $c_t^s,a_t^s\in[0,1]$. A larger $c_t^s$ indicates a higher-certainty prediction state and a larger $a_t^s$ indicates greater ambiguity.

We then define the token decision interval margin with $c_t^s$:
\begin{equation}
m_t^s = \theta^{s}_{u} - \theta^{s}_{d} - c^{s}_{t}.
\label{eq:interval_margin}
\end{equation}
A token is retained when $m_t^s\geq0$, which indicates an unresolved token. Otherwise, the token is considered sufficiently resolved and is removed from subsequent computation. 


Based on the interval margin, we construct a binary decision mask $M^s$ over the $N_{s-1}$ active patch tokens:
\begin{equation}
M^s\in\{0,1\}^{N_{s-1}\times1},\qquad
M^s(t)=
\begin{cases}
1, & m_t^s\geq0,\\
0, & m_t^s<0.
\end{cases}
\label{eq:binary_mask}
\end{equation}
The mask is progressively updated across pruning stages as token representations evolve. Consequently, the number of active patch tokens retained changes to $N_s$.

Since COD relies on complete spatial representations, the sparsified features need to be restored by token refilling. Let $\mathbf{F}^s \in \mathbb{R}^{N_s \times C}$ denote the sparse features at stage $s$. For each token removed, we cache its last representation together with its original patch index. Before decoding, the cached tokens and the remaining active tokens are scattered back to their original spatial locations, reconstructing a complete feature map $\hat{\mathbf{F}}^s \in \mathbb{R}^{T \times C}$ over $T$ spatial positions. 
\subsection{Dual-Path Feature Compensation}
\label{sec:compensation}
According to the former description, the active token set $\mathcal{T}_s$ is divided into
\begin{equation}
\left\{
\begin{aligned}
\mathcal{U}^{s}_{mid} &= \left\{t\in\mathcal{T}_s\mid m_t^s\geq0\right\},\\
\mathcal{U}^{s}_{low} &= \left\{t\in\mathcal{T}_s\mid m_t^s<0,\ p_t^s<\frac{1}{2} \right\},\\
\mathcal{U}^{s}_{high} &= \left\{t\in\mathcal{T}_s\mid m_t^s<0,\ p_t^s>
\frac{1}{2} \right\}.
\end{aligned}
\right.
\label{eq:token_partition}
\end{equation}
Here, $\mathcal{U}^{s}_{mid}$ contains low-certainty ambiguous tokens, while $\mathcal{U}^{s}_{low}$ and $\mathcal{U}^{s}_{high}$ contain high-certainty background and foreground tokens, respectively.

Although high-certainty foreground and background tokens are removed from subsequent token-wise computation, their features may still contain contextual cues useful for distinguishing camouflaged objects from their surroundings. To retain such information without restoring the original token sequence, we introduce a \textbf{D}ual-\textbf{P}ath \textbf{F}eature \textbf{C}ompensation (\textbf{DPFC}) mechanism. Specifically, token sets $\mathcal{U}^{s}_{low}$ and $\mathcal{U}^{s}_{high}$ are separately compressed into two compact prototype tokens $\mathbf{z}^{s}_{low},\mathbf{z}^{s}_{high}\in\mathbb{R}^{1\times C}$ through aggregation:
\begin{equation}
\begin{aligned}
\mathbf{z}^{s}_{\mathrm{low}}
&=
\sum_{t\in\mathcal{U}^{s}_{\mathrm{low}}}
\frac{p_t^s}
{\sum_{k\in\mathcal{U}^{s}_{\mathrm{low}}}p_k^s}
\mathbf{x}_t,
\\
\mathbf{z}^{s}_{\mathrm{high}}
&=
\sum_{t\in\mathcal{U}^{s}_{\mathrm{high}}}
\frac{1-p_t^s}
{\sum_{k\in\mathcal{U}^{s}_{\mathrm{high}}}(1-p_k^s)}
\mathbf{x}_t,
\end{aligned}
\label{eq:prototype_aggregation}
\end{equation}
where $\mathbf{x}_t\in\mathbb{R}^{1\times C}$ denotes the feature vector of token $t$. Within the background group $\mathcal{U}^{s}_{low}$, a larger $p_t^s$ indicates that the token is relatively closer to the retained ambiguity interval and therefore assigns it a larger aggregation weight. Similarly, within the foreground group $\mathcal{U}^{s}_{high}$, a larger $1-p_t^s$ gives greater weight to tokens closer to the ambiguity interval. Thus, the weighting strategy emphasizes relatively more ambiguous tokens within each high-certainty group, allowing the two prototypes to preserve informative context from the removed token sets.

Following aggregation, the input sequence for the subsequent Transformer stage is constructed by concatenating the class token, the retained ambiguous patch tokens, and the two prototype tokens:
\begin{equation}
\tilde{\mathbf{X}}^{s}=[\mathbf{x}_{cls};\ \mathbf{X}_{mid};\ \mathbf{z}^{s}_{low};\ \mathbf{z}^{s}_{high}]\in\mathbb{R}^{(N_{mid}+3)\times C},
\label{eq:reconstructed_sequence}
\end{equation}
where $\mathbf{X}_{mid}$ denotes the features of the retained tokens in $\mathcal{U}^{s}_{mid}$, and $N_{mid}=|\mathcal{U}^{s}_{mid}|$. The two prototype tokens is removed from the token sequence after the network layer computation in this stage is completed. Empty prototypes are omitted from concatenation. Accordingly, the sequence length varies with the number of active token groups.

\begin{figure*}[!t]
    \centering 
    \includegraphics[width=.85\textwidth]{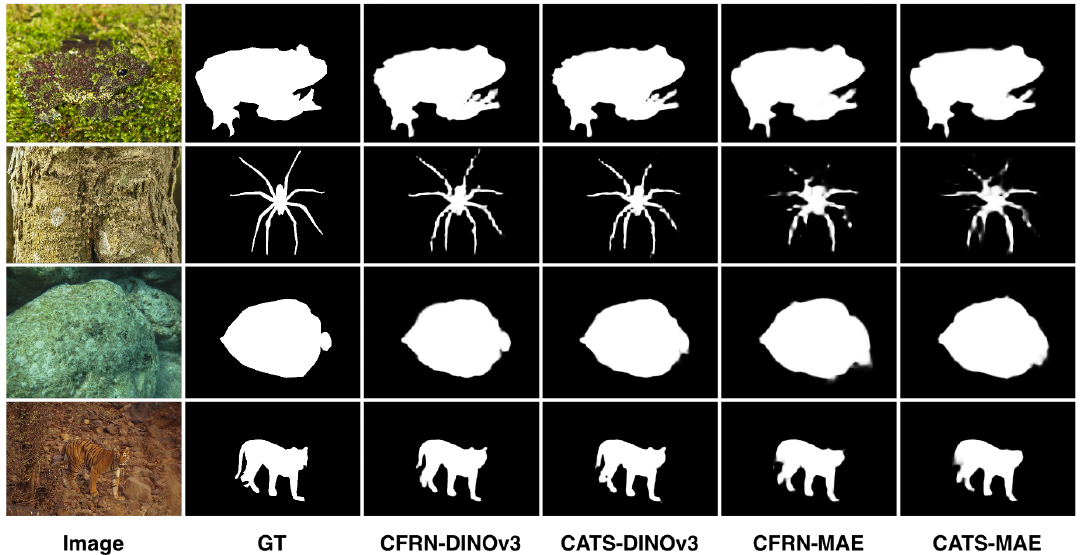} 
    \caption{Qualitative comparisons on four standard COD datasets. From left to right: input images, ground-truth (GT) masks, CFRN-DINOv3 predictions w/o and w/ CATS, and CFRN-MAE predictions w/o and w/ CATS.}
    \label{fig:compare_vis} 
\end{figure*}

\section{Experiments}
\subsection{Experimental Setup}
\subsubsection{Datasets.}
We conduct comprehensive evaluation on four widely used COD benchmarks: CAMO~\cite{le2019anabranch}, CHAMELEON~\cite{skurowski2018animal}, COD10K~\cite{fan2020camouflaged}, and NC4K~\cite{lv2021simultaneously}. \textbf{CAMO} comprises 1,250 images across 8 categories, encompassing both natural and artificial camouflage scenarios. \textbf{CHAMELEON} provides a high-resolution collection of 76 meticulously annotated images. \textbf{COD10K} contains 5,066 diverse images curated from various photography platforms to ensure high semantic variety. \textbf{NC4K} consists of 4,121 images and is primarily utilized to assess the generalization and robustness of COD models. Following the standard protocol, we merge CAMO and COD10K for training, use CHAMELEON for validation, and evaluate on the test sets of all four datasets.

\begin{table*}[!t]
  \centering
  \renewcommand{\arraystretch}{1.3}
  \resizebox{\textwidth}{!}{
  \begin{tabular}{c cc cccc cccc cccc cccc}
  \toprule
  \multirow{2}{*}{\textbf{Sparsification Strategy}} & \multirow{2}{*}{\textbf{GFLOPs} $\downarrow$} & \multirow{2}{*}{\textbf{FPS} $\uparrow$} 
  & \multicolumn{4}{c}{\textbf{CAMO}} 
  & \multicolumn{4}{c}{\textbf{CHAMELEON}} 
  & \multicolumn{4}{c}{\textbf{COD10K}} 
  & \multicolumn{4}{c}{\textbf{NC4K}} \\
  \cmidrule(lr){4-7} \cmidrule(lr){8-11} \cmidrule(lr){12-15} \cmidrule(lr){16-19} 
   & & 
   & $S_\alpha \uparrow$ & $E_\xi \uparrow$ & $F_\beta^w \uparrow$ & $\mathcal{M} \downarrow$ & $S_\alpha \uparrow$ & $E_\xi \uparrow$ & $F_\beta^w \uparrow$ & $\mathcal{M} \downarrow$ & $S_\alpha \uparrow$ & $E_\xi \uparrow$ & $F_\beta^w \uparrow$ & $\mathcal{M} \downarrow$ & $S_\alpha \uparrow$ & $E_\xi \uparrow$ & $F_\beta^w \uparrow$ & $\mathcal{M} \downarrow$ \\
  \midrule

  CFRN & 359.6 & 33.1
  & .915 & .955 & .890 & .029
  & .932 & .972 & .902 & .017
  & .912 & .963 & .860 & .015
  & .922 & .961 & .895 & .022 \\

  \cmidrule(lr){1-19}

  $\text{CATS}_{\text{certain}}$ & 237.8 & 44.4 
  & .911 & .954 & .887 & .030
  & .929 & .971 & .900 & .017 
  & .903 & .957 & .848 & .016
  & .914 & .955 & .884 & .024 \\

  $\text{CATS}_{\text{uncertain}}$ & 146.1 & 61.4 
  & .731 & .782 & .590 & .110
  & .806 & .865 & .671 & .061 
  & .742 & .810 & .542 & .058
  & .771 & .821 & .634 & .081 \\
  \cmidrule(lr){1-19}
    
  DynamicViT & 237.8 & 46.1 
  & .642 & .548 & .335 & .184
  & .702 & .616 & .368 & .142 
  & .647 & .584 & .263 & .126
  & .678 & .595 & .354 & .154 \\ 

  ToMe & 223.0 & 44.7 
  & .882 & .941 & .843 & .039
  & .878 & .941 & .812 & .027 
  & .868 & .947 & .789 & .022
  & .890 & .948 & .848 & .029 \\

  DToP & 240.1 & 38.2
  & .907 & .940 & .862 & .035
  & .924 & .959 & .870 & .021
  & .892 & .936 & .800 & .021
  & .905 & .938 & .851 & .030\\

  DAR-TR-PEFT & 301.8 & 30.1 
  & .911 & .951 & .884 & .032
  & .933 & .972 & .901 & .017
  & .907 & .959 & .850 & .017
  & .917 & .955 & .885 & .026 \\
  \bottomrule
  \end{tabular}}
  \caption{Validation of certain tokens' redundancy and performance comparison of different sparsification strategies transferred to COD on $\text{CFRN}_{\text{DINOv3}}$-base baseline.}
  \label{tab:certainty}
\end{table*}

\subsubsection{Implementation Details.}
All experiments are conducted on a single NVIDIA RTX 4090 GPU. We utilize DINOv3 and MAE as the backbone using 768 and 512 as image size respectively and follow their official network parameters. Unless otherwise specified, the certainty thresholds are set to $0.3/0.7$ which remain constant across all stages and for the most representitive base scale setting, we utilize $S=4$ stages. To ensure a fair comparison, our training configurations strictly adhere to the original baseline settings and parameters on two SoTA COD baselines CFRN~\cite{song2025continuous} and SENet~\cite{hao2025simple}. Specifically, we retain each baseline's standard loss function and additionally supervise each learnable CATS prediction head using its primary prediction loss, with an auxiliary-loss weight of 0.3.

\subsubsection{Evaluation Metrics.} 
To quantitatively assess segmentation performance, we adopt four widely recognized COD metrics: Structure-measure ($S_{\alpha}$) \cite{fan2017structure}, weighted F-measure ($F_{\beta}^{w}$) \cite{margolin2014evaluate}, mean E-measure ($E_{\xi}$) \cite{fan2018enhanced} and mean absolute error ($\mathcal{M}$)~\cite{perazzi2012saliency}. Furthermore, to evaluate the efficiency of our token pruning strategy, we report GFLOPs and Frames Per Second (FPS) as key indicators of computational complexity and inference speed, respectively.

\subsection{Ablation Studies}
\begin{table*}[!t]
  \centering
  \renewcommand{\arraystretch}{1.3}
  \resizebox{\textwidth}{!}{
  \begin{tabular}{c cc cccc cccc cccc cccc}
  \toprule
  \multirow{2}{*}{\textbf{THresholds and Stage}} & \multirow{2}{*}{\textbf{GFLOPs} $\downarrow$} & \multirow{2}{*}{\textbf{FPS} $\uparrow$} 
  & \multicolumn{4}{c}{\textbf{CAMO}} 
  & \multicolumn{4}{c}{\textbf{CHAMELEON}} 
  & \multicolumn{4}{c}{\textbf{COD10K}} 
  & \multicolumn{4}{c}{\textbf{NC4K}} \\
  \cmidrule(lr){4-7} \cmidrule(lr){8-11} \cmidrule(lr){12-15} \cmidrule(lr){16-19} 
   & & 
   & $S_\alpha \uparrow$ & $E_\xi \uparrow$ & $F_\beta^w \uparrow$ & $\mathcal{M} \downarrow$
   & $S_\alpha \uparrow$ & $E_\xi \uparrow$ & $F_\beta^w \uparrow$ & $\mathcal{M} \downarrow$
   & $S_\alpha \uparrow$ & $E_\xi \uparrow$ & $F_\beta^w \uparrow$ & $\mathcal{M} \downarrow$
   & $S_\alpha \uparrow$ & $E_\xi \uparrow$ & $F_\beta^w \uparrow$ & $\mathcal{M} \downarrow$ \\
  \midrule


  
  0.4/0.6 & 331.3 & 33.3 
  & .913 & .953 & .890 & .030
  & .932 & .971 & .905 & .017
  & .909 & .962 & .858 & .016
  & .917 & .957 & .889 & .024 \\ 

  0.35/0.65 & 279.6 & 38.9 
  & .912 & .954 & .890 & .030
  & .932 & .972 & .901 & .017
  & .907 & .960 & .854 & .016
  & .916 & .956 & .887 & .024 \\ 

  0.3/0.7 & 237.8 & 44.4 
  & .911 & .954 & .887 & .030
  & .929 & .971 & .900 & .017 
  & .903 & .957 & .848 & .016
  & .914 & .955 & .884 & .024 \\
    
  0.25/0.75 & 202.4 & 50.1 
  & .906 & .948 & .877 & .033
  & .926 & .969 & .893 & .018
  & .897 & .954 & .836 & .018
  & .908 & .950 & .874 & .027 \\ 

  0.2/0.8 & 173.4 & 57.7 
  & .892 & .941 & .854 & .039
  & .912 & .959 & .869 & .023
  & .879 & .940 & .803 & .023
  & .895 & .938 & .850 & .033 \\ 

    \cmidrule(lr){1-19}

  layer 3 & 345.5 & 33.6 
  & .912 & .955 & .889 & .029
  & .931 & .971 & .902 & .017
  & .912 & .964 & .861 & .015
  & .920 & .960 & .893 & .022 \\

  layer 3, 6 & 248.2 & 43.6 
  & .912 & .951 & .885 & .031
  & .930 & .971 & .898 & .017
  & .905 & .960 & .851 & .016
  & .916 & .956 & .886 & .024 \\

  layer 3, 6, 9 & 237.8 & 44.4 
  & .911 & .954 & .887 & .030
  & .929 & .971 & .900 & .017 
  & .903 & .957 & .848 & .016
  & .914 & .955 & .884 & .024 \\
  \bottomrule
  \end{tabular}}
  \caption{Performance comparison of CATS certainty thresholds and sparsification stages on the $\text{CFRN}_{\text{DINOv3}}$-base baseline.}
  \label{tab:thresholds}
  \vspace{-3.5mm}
\end{table*}

\begin{table*}[!t]
  \centering
  \renewcommand{\arraystretch}{1.3}
  \resizebox{\textwidth}{!}{
  \begin{tabular}{c cc cccc cccc cccc cccc} 
  \toprule
  \multirow{2}{*}{\textbf{Aggregation Setting}} & \multirow{2}{*}{\textbf{GFLOPs} $\downarrow$} & \multirow{2}{*}{\textbf{FPS} $\uparrow$} 
  & \multicolumn{4}{c}{\textbf{CAMO}} & \multicolumn{4}{c}{\textbf{CHAMELEON}} 
  & \multicolumn{4}{c}{\textbf{COD10K}} & \multicolumn{4}{c}{\textbf{NC4K}} \\
  \cmidrule(lr){4-7} \cmidrule(lr){8-11} \cmidrule(lr){12-15} \cmidrule(lr){16-19} 
   & & & $S_\alpha \uparrow$ & $E_\xi \uparrow$ & $F_\beta^w \uparrow$ & $\mathcal{M} \downarrow$ & $S_\alpha \uparrow$ & $E_\xi \uparrow$ & $F_\beta^w \uparrow$ & $\mathcal{M} \downarrow$ & $S_\alpha \uparrow$ & $E_\xi \uparrow$ & $F_\beta^w \uparrow$ & $\mathcal{M} \downarrow$ & $S_\alpha \uparrow$ & $E_\xi \uparrow$ & $F_\beta^w \uparrow$ & $\mathcal{M} \downarrow$ \\
  \midrule

  \ding{56} & 270.0 & 39.8 
  & .905 & .947 & .876 & .033
  & .929 & .969 & .897 & .018 
  & .898 & .953 & .836 & .017
  & .910 & .951 & .876 & .026 \\

  Average & 297.5 & 37.7 
  & .910 & .951 & .885 & .032
  & .931 & .971 & .900 & .016 
  & .906 & .958 & .850 & .016
  & .914 & .954 & .882 & .024 \\  
  
  DPFC & 237.8 & 44.4 
  & .911 & .954 & .887 & .030
  & .929 & .971 & .900 & .017 
  & .903 & .957 & .848 & .016
  & .914 & .955 & .884 & .024 \\
  \bottomrule
  \end{tabular}}
  \caption{Performance comparison of different feature compensation strategies with the same CATS certainty threshold on the $\text{CFRN}_{\text{DINOv3}}$-base baseline.}
  \label{tab:abla_compensation}
\end{table*}

\subsubsection{Effect of Token Certainty.}
To validate the core insight that ``certainty is redundant'', we conduct comparative experiments using two different token sparsification criteria. One group involves pruning tokens identified with high certainty, while the other involves the opposite approach: pruning uncertain tokens. The results are presented in the upper section of \cref{tab:certainty}. Compared to the CFRN baseline \cite{song2025continuous}, CATS-uncertain achieves a 59.4\% reduction in GFLOPs and a 85.5\% increase in FPS, but suffers a significant accuracy drop in $S_\alpha$, $E_\xi$, $F_\beta^w$, $\mathcal{M}$ for 15.7\%, 14.3\%, 19.7\%, 5.7\% on average, respectively. In contrast, CATS-certain achieves a 33.9\% reduction in GFLOPs and a 34.1\% increase in FPS while maintaining accuracy. Furthermore, results in the below section of \cref{tab:thresholds} also show that removing certain tokens does not significantly compromise model accuracy, yet effectively reduces overall computational load and accelerates inference. 
These findings validate our insight, certainty is redundant, is valid and effective.
\begin{table*}[!t]
  \centering
  \renewcommand{\arraystretch}{1.3}
  \resizebox{\textwidth}{!}{
  \begin{tabular}{ccc cccc cccc cccc cccc} 
  \toprule
  \multirow{2}{*}{\textbf{Method}} & \multirow{2}{*}{\textbf{GFLOPs} $\downarrow$} & \multirow{2}{*}{\textbf{FPS} $\uparrow$} & \multicolumn{4}{c}{\textbf{CAMO}} & \multicolumn{4}{c}{\textbf{CHAMELEON}} & \multicolumn{4}{c}{\textbf{COD10K}} & \multicolumn{4}{c}{\textbf{NC4K}} \\
  \cmidrule(lr){4-7} \cmidrule(lr){8-11} \cmidrule(lr){12-15} \cmidrule(lr){16-19} 
   & & & $S_\alpha \uparrow$ & $E_\xi \uparrow$ & $F_\beta^w \uparrow$ & $\mathcal{M} \downarrow$ & $S_\alpha \uparrow$ & $E_\xi \uparrow$ & $F_\beta^w \uparrow$ & $\mathcal{M} \downarrow$ & $S_\alpha \uparrow$ & $E_\xi \uparrow$ & $F_\beta^w \uparrow$ & $\mathcal{M} \downarrow$ & $S_\alpha \uparrow$ & $E_\xi \uparrow$ & $F_\beta^w \uparrow$ & $\mathcal{M} \downarrow$ \\
  \midrule

    $\text{CFRN}_{\text{DINOv3}}$ & 359.6 & 33.1
  & .915 & .955 & .890 & .029
  & .932 & .972 & .902 & .017
  & .912 & .963 & .860 & .015
  & .922 & .961 & .895 & .022 \\
  
  $\text{CFRN}_{\text{CATS}}$ & 237.8 & 45.5 
  & .911 & .954 & .887 & .030
  & .929 & .971 & .900 & .017 
  & .903 & .957 & .848 & .016
  & .914 & .955 & .884 & .024 \\
    \cmidrule(lr){1-19}
\textit{$\Delta$} & \textbf{$-33.9\%$} & +12.4 & 
+0.4\% & +0.5\% & +0.0\% & +0.1\% & 
-0.2\% & -0.7\% & -0.1\% & +0.1\% & 
-0.7\% & -1.3\% & -0.5\% & +0.2\% & 
-0.6\% & -0.9\% & -0.5\% & +0.2\% \\

  \cmidrule(lr){1-19}

  $\text{CFRN}_{\text{MAE}}$ & 135.4 & 68.7
  & .879 & .927 & .836 & .043
  & .899 & .955 & .846 & .023
  & .864 & .931 & .775 & .024
  & .888 & .936 & .840 & .032 \\
  
  $\text{CFRN}_{\text{CATS}}$ & 97.8 & 77.3 
  & .876 & .925 & .835 & .043
  & .906 & .959 & .855 & .022
  & .861 & .929 & .774 & .025
  & .884 & .933 & .835 & .034\\
    \cmidrule(lr){1-19}
\textit{$\Delta$} & \textbf{$-27.8\%$} & +8.6 & 
-0.3\% & -0.2\% & -0.1\% & - & 
+0.7\% & +0.4\% & +0.9\% & -0.1\% & 
-0.3\% & -0.2\% & -0.1\% & +0.1\% & 
-0.4\% & -0.3\% & -0.5\% & +0.2\% \\

   \cmidrule(lr){1-19}
  $\text{SENet}_{\text{MAE}}$ & 142.3 & 40.4 &
  .868 & .908 & .821 & .048 &
  .907 & .943 & .862 & .025 &
  .861 & .913 & .769 & .026 &
  .882 & .921 & .832 & .036 \\ 
  
  $\text{SENet}_{\text{CATS}}$ & 99.7 & 44.0 &
  .858 & .900 & .814 & .051 &
  .904 & .944 & .858 & .023 &
  .855 & .912 & .767 & .028 &
  .874 & .915 & .826 & .039\\
  \cmidrule(lr){1-19} 
\textit{$\Delta$} & \textbf{$-29.9\%$}& +3.6 & 
-1.0\% & -0.8\% & -0.7\% & +0.3\% & 
-0.3\% & +0.1\% & -0.4\% & -0.2\% & 
-0.6\% & -0.1\% & -0.2\% & +0.2\% & 
-0.8\% & -0.6\% & -0.6\% & +0.3\% \\

  \bottomrule
  \end{tabular}}
  \caption{Performance comparison on different VFMs and SoTA COD baselines w/ and w/o CATS.}
  \label{tab:results_sota}
\end{table*}

\begin{table*}[!t]
  \centering
  \renewcommand{\arraystretch}{1.3}
  \resizebox{\textwidth}{!}{
  \begin{tabular}{ccc cccc cccc cccc cccc} 
  \toprule
  \multirow{2}{*}{\textbf{Backbone Scale}} & \multirow{2}{*}{\textbf{GFLOPs} $\downarrow$} & \multirow{2}{*}{\textbf{FPS} $\uparrow$}  & \multicolumn{4}{c}{\textbf{CAMO}} & \multicolumn{4}{c}{\textbf{CHAMELEON}} & \multicolumn{4}{c}{\textbf{COD10K}} & \multicolumn{4}{c}{\textbf{NC4K}} \\
  \cmidrule(lr){4-7} \cmidrule(lr){8-11} \cmidrule(lr){12-15} \cmidrule(lr){16-19} 
   & & & $S_\alpha \uparrow$ & $E_\xi \uparrow$ & $F_\beta^w \uparrow$ & $\mathcal{M} \downarrow$ & $S_\alpha \uparrow$ & $E_\xi \uparrow$ & $F_\beta^w \uparrow$ & $\mathcal{M} \downarrow$ & $S_\alpha \uparrow$ & $E_\xi \uparrow$ & $F_\beta^w \uparrow$ & $\mathcal{M} \downarrow$ & $S_\alpha \uparrow$ & $E_\xi \uparrow$ & $F_\beta^w \uparrow$ & $\mathcal{M} \downarrow$ \\
   \cmidrule(lr){1-19}
  Small & 115.4 & 72.3
  & .903 & .948 & .872 & .034
  & .926 & .969 & .889 & .018
  & .898 & .954 & .835 & .018
  & .913 & .954 & .880 & .024 \\
  
  $\text{Small}_{\text{CATS}}$ & 83.0 & 77.43 &
  .902 & .947 & .871 & .035 &
  .921 & .967 & .884 & .019 &
  .894 & .952 & .831 & .018 &
  .907 & .949 & .871 & .026 \\

  Base & 359.6 & 33.1
  & .915 & .955 & .890 & .029
  & .932 & .972 & .902 & .017
  & .912 & .963 & .860 & .015
  & .922 & .961 & .895 & .022 \\

  $\text{Base}_{\text{CATS}}$ & 237.8 & 44.4 
  & .911 & .954 & .887 & .030
  & .929 & .971 & .900 & .017 
  & .903 & .957 & .848 & .016
  & .914 & .955 & .884 & .024 \\
  
  Large & 1073.9 & 12.0
  & .921 & .958 & .900 & .027
  & .936 & .974 & .912 & .016
  & .921 & .970 & .876 & .014
  & .926 & .963 & .903 & .021 \\
  
  $\text{Large}_{\text{CATS}}$ & 650.8 & 19.1 &
  .922 & .963 & .904 & .026 &
  .935 & .972 & .908 & .016 &
  .914 & .965 & .866 & .014 &
  .920 & .958 & .895 & .022 \\

  Huge & 2551.6 & 5.9 &
  .924 & .961 & .906 & .025 &
  .938 & .975 & .914 & .016 &
  .924 & .971 & .882 & .013 &
  .929 & .967 & .909 & .019 \\

  $\text{Huge}_{\text{CATS}}$ & 1247.8 & 11.5 &
  .919 & .959 & .899 & .028 &
  .935 & .971 & .907 & .017 &
  .915 & .966 & .867 & .014 &
  .922 & .960 & .898 & .022 \\
  \bottomrule
  \end{tabular}}
  \caption{Performance comparison of different backbone scales on the $\text{CFRN}_{\text{DINOv3}}$ baseline w/ and w/o CATS.}
  \label{tab:scale}
  \vspace{-3.5mm}
\end{table*}

\subsubsection{Comparison with Other Sparsification Methods.}
To fully validate the effectiveness of our proposed CATS in VFM-based COD, we adapt DynamicViT, ToMe, DToP, and DAR-TR-PEFT to the same COD framework using the best-performing hyperparameter configurations they reported, while keeping the remaining training settings unchanged. The results are shown in the lower half of \cref{tab:certainty} and can be compared with \cref{tab:thresholds}. The transfer results of DynamicViT and ToMe shows that although its flops and FPS are similar to our CATS, its accuracy differs significantly from the baseline. These methods originally developed primarily for image-level recognition and are not robust to directly adapted to COD under comparable computational budgets in our experiments. As for the adaption of DToP and DAR-TR-PEFT, the results indicates that when the accuracy of DAR-TR-PEFT is comparable to our CATS, its GFLOPs and FPS lag behind ours. Conversely, when DToP’s GFLOPs and FPS levels are close to our CATS, its accuracy falls short of ours.
This proves that our CATS is more suitable for COD.

\subsubsection{Effect of Certainty Thresholds and Stage Settings.}
We vary threshold $\theta_{d}$ and $\theta_{u}$ and sparsification stages to examine the sensitivity of determine an appropriate accuracy–efficiency trade-off. The results are presented in \cref{tab:thresholds}. After comparing results across multiple threshold combinations and stage settings, the 0.3/0.7 threshold combination appears to be the optimal choice. The 0.4/0.6 reults achieves significant improvements in FLOPs and FPS yet it shows a slight accuracy drop. While the 0.2/0.8 setting results shows the most efficiency gain, it fails to maintain competitive accuracy levels relative to the baseline. Comparisons among sparsification stages {3}, {3,6}, and {3,6,9} indicate that staged pruning, particularly the {3,6,9} configuration, leads to a better accuracy–efficiency trade-off than one-shot pruning.

\subsubsection{Effect of Dual-Path Feature Compensation.}
To validate the effectiveness of our proposed DPFC, we set up comparative experiments evaluating three settings all along with same CATS: no aggregation, average aggregation, and DPFC. Results in \cref{tab:abla_compensation} show that, compared to the no-aggregation and average aggregation settings, DPFC further reduces computational load and increases inference speed while achieving a light accuracy improvement. These experiments and analyses demonstrate the effectiveness of our DPFC method. Although the same certainty threshold is used, different compensation strategies affect subsequent token representations and prediction certainty, resulting in different numbers of retained tokens and therefore different GFLOPs.

\subsubsection{Effect of Backbone Scale.}
To fully demonstrate the generalization capability of our proposed CATS and verify the performance gains of VFMs on the COD task, we present CATS results on multiple DINOv3 backbone scales in \cref{tab:scale}. Across the various backbone scales tested, our CATS method maintains pre-pruning accuracy levels while achieving a 28.1\%–51.1\% reduction in FLOPs and up to 94.9\% improvement in FPS. This demonstrates that VFMs, leveraging exceptional general-purpose visual features learned from large-scale datasets, can successfully transfer to the COD task and achieve competitive results. Furthermore, CATS strikes an effective balance between efficiency and accuracy across multiple backbone scales.

\subsection{Comparative Pruning Results on SoTA Models}
\cref{tab:results_sota} reports the main results on three representative SoTA models, \ie, SENet~\cite{hao2025simple}, and CFRN~\cite{song2025continuous}. 
Overall, our CATS achieves substantial computational reduction and inference acceleration while maintaining competitive performance across all four benchmark datasets.

For CFRN-DINOv3, CATS achieves 33.9\% reduction and 12.4 FPS improvement in GFLOPs and FPS with only a small accuracy drop in $S_\alpha$, $E_\xi$, $F_\beta^w$, $\mathcal{M}$ for 0.6\%, 0.4\%, 0.7\%, 0.1\% on average respectively.
For SENet, CATS reduces the computational cost by 29.4\% GFLOPs and 3.6 in FPS with the performance degradation remains limited in $S_\alpha$, $E_\xi$, $F_\beta^w$, $\mathcal{M}$ for 0.7\%, 0.3\%, 0.5\%, 0.1\% on average respectively. 
For CFRN, the efficiency gain 27.8\% in GFLOPs and 8.6 in speed on MAE. Importantly, the accuracy only reduces no more than 1.3\%.
Qualitative comparisons in \cref{fig:compare_vis} further demonstrate that predictions from pruned models remain highly consistent with ground truth and the progressive pruning process is visualized in \cref{fig:heat map} and \cref{fig:certainty}.

\section{Conclusion}
This work presents a selective computation allocation in VFM-based COD. Experiments reveal that certain tokens with stabilized prediction states is redundant for further refinement. CATS exploits this by progressively removing them from subsequent updates, while DPFC retains their contextual value. Across multiple COD benchmarks, VFMs, and COD baselines, CATS reduces FLOPs and improves inference speed with only marginal accuracy changes. These results indicate that our formulation, certainty is redundant, is crucial, CATS and DPFC effectively sparse superfluous tokens with a comparable accuracy mainteinance helping achieve Efficient VFMs-based COD.
Future work will explore hardware-aware deployment and integration with complementary efficiency techniques, while extending certainty-guided token sparsification to other vision tasks.

\section*{Acknowledgements}
This work was supported by the National Natural Science Foundation of China (No. 62576176). The computational resources were supported by the Supercomputing Center of Nankai University (NKSC).

\bibliography{aaai2027}


\end{document}